\documentclass[letterpaper, 10 pt, conference]{ieeeconf}  % Comment this line out if you need a4paper

\IEEEoverridecommandlockouts                              % This command is only needed if 
\usepackage[utf8]{inputenc}
\usepackage{graphicx}
\usepackage{amssymb}
\usepackage{amsmath}
\usepackage{xcolor}

\usepackage{booktabs} 
\usepackage{caption} %
\usepackage{algorithm}
\usepackage{algorithmic}
\usepackage{mdframed}
\usepackage{verbatim}
\usepackage{cite}
\usepackage{authblk}
\usepackage[font=footnotesize]{caption}
\definecolor{deepblue}{rgb}{0.1, 0.3, 0.6}   % 深蓝色
\definecolor{lightgreen}{HTML}{66CC66}
\title{\LARGE \bf
Learning Task-Invariant Properties via Dreamer: Enabling Efficient Policy Transfer for Quadruped Robots
}

\author{\textbf{Junyang Liang$^{1}$, Yuxuan Liu$^{1}$, Yabin Chang$^{1}$, Junfan Lin$^{2}$, Junkai Ji$^{1}$, Hui Li$^{1}$, } \\
\textbf{Changxin Huang$^{1*}$, Jianqiang Li$^{1}$}
\thanks{*Corresponding Author: Changxin Huang (huangchx@szu.edu.cn)}
\thanks{\textsuperscript{1}{Shenzhen University, Shenzhen, China}}
\thanks{\textsuperscript{2}{Peng Cheng Laboratory, Shenzhen, China}}
\thanks{This work is supported in part by the National Natural Science Foundation of China (No. 62403325, No. 62325307, No. 62527809, No. 62203134, No. 62373258, No. 62506180), in part by the Natural Science Foundation of Guangdong Province (No. 2023B1515120038, No. 2026A1515011532), in part by Shenzhen Science and Technology Innovation Commission (No. 20231122104038002, No. KJZD20230923113801004, No. JCYJ20240813141628038, No. KJZD20230923115215032), in part by the Shenzhen Key Industry R\&D Program Project (No. ZDCY20250901102300001), in part by China Postdoctoral Science Foundation (No. 2025M771522), in part by the Major Key Project of PCL (No. PCL2024A04, No. PCL2025A17). This work is also supported by the Intelligent Computing Center of Shenzhen University.}% <-this % stops a space
}

\begin{document}

\maketitle{}
\thispagestyle{empty}
\pagestyle{empty}

%%%%%%%%%%%%%%%%%%%%%%%%%%%%%%%%%%%%%%%%%%%%%%%%%%%%%%%%%%%%%%%%%%%%%%%%%%%%%%%%
\begin{abstract}
%在多样且动态的地形上实现四足机器人运动具有显著挑战，主要原因在于仿真环境与真实环境之间存在差异。而传统的虚实迁移方法则依赖人工特征设计或高成本的真实环境微调。为解决这些局限，本文提出 DreamTIP 框架，该方法在 Dreamer 世界模型架构中引入任务不变性属性学习，以增强虚实迁移能力。借助大语言模型的指导，DreamTIP 能够设计如接触稳定性、地形间隙等对动力学变化鲁棒且可跨任务迁移的属性，并将其作为辅助预测目标融入世界模型，从而使策略学习到对底层动力学变化不敏感的表征。此外，本文设计了一种高效的适应策略，通过混合回放缓冲池与正则化约束，实现对真实动力学的快速校准，同时有效避免表征崩塌和灾难性遗忘。在包括楼梯、攀爬和爬行等复杂地形上的大量实验表明，DreamTIP 在仿真和真实环境中均显著优于现有最先进基线。结果表明，将任务不变性属性引入世界模型学习，为实现鲁棒、具备迁移能力的机器人运动提供了一种新颖的解决方案。
Achieving quadruped robot locomotion across diverse and dynamic terrains presents significant challenges, primarily due to the discrepancies between simulation environments and real-world conditions. Traditional sim-to-real transfer methods often rely on manual feature design or costly real-world fine-tuning. To address these limitations, this paper proposes the DreamTIP framework, which incorporates Task-Invariant Properties learning within the Dreamer world model architecture to enhance sim-to-real transfer capabilities. Guided by large language models, DreamTIP identifies and leverages Task-Invariant Properties, such as contact stability and terrain clearance, which exhibit robustness to dynamic variations and strong transferability across tasks. These properties are integrated into the world model as auxiliary prediction targets, enabling the policy to learn representations that are insensitive to underlying dynamic changes. Furthermore, an efficient adaptation strategy is designed, employing a mixed replay buffer and regularization constraints to rapidly calibrate to real-world dynamics while effectively mitigating representation collapse and catastrophic forgetting. Extensive experiments on complex terrains, including Stair, Climb, Tilt, and Crawl, demonstrate that DreamTIP significantly outperforms state-of-the-art baselines in both simulated and real-world environments. Our method achieves an average performance improvement of 28.1\% across eight distinct simulated transfer tasks. In the real-world Climb task, the baseline method achieved only a 10\% success rate, whereas our method attained a 100\% success rate. These results indicate that incorporating Task-Invariant Properties into Dreamer learning offers a novel solution for achieving robust and transferable robot locomotion.
\end{abstract}

%%%%%%%%%%%%%%%%%%%%%%%%%%%%%%%%%%%%%%%%%%%%%%%%%%%%%%%%%%%%%%%%%%%%%%%%%%%%%%%%
\section{INTRODUCTION}

%让机器人在复杂非结构化的真实世界中自主完成任务是具身智能的终极目标\cite{kim2025high}。然而，物理实体训练存在数据效率低、试错成本高与安全风险大等问题\cite{10160497,long2023hybrid}，因此利用仿真环境中低成本、大规模的数据进行训练已成为广泛采用的范式\cite{zhao2020sim}。尽管如此，仿真与真实环境间存在固有的动力学差异（dynamics gap），以及传感器噪声和视觉不匹配等问题，导致基于仿真数据训练的策略在真实环境中部署时普遍出现性能下降\cite{wu2023learning,cheng2024extreme}。因此，如何实现仿真训练智能体向真实世界的高效、稳定迁移，并在未知环境中保持泛化能力，构成了 Sim-to-Real 迁移的核心挑战\cite{he2025attention,zhang2024learning,kim2025stage}。
Enabling robots to autonomously operate in complex real-world environments remains a core goal of embodied intelligence \cite{kim2025high}. However, real-world training faces challenges such as low data efficiency, high costs, and safety risks \cite{10160497, long2023hybrid}. Training in simulation offers a low-cost, scalable alternative \cite{zhao2020sim} , yet dynamics discrepancies, sensor noise, and visual mismatches often cause performance drops when transferring policies to reality \cite{wu2023learning, cheng2024extreme}. Thus, achieving efficient and stable sim-to-real transfer while maintaining generalization in unseen environments is the key challenge \cite{he2025attention, zhang2024learning, kim2025stage}.
\begin{figure}[t]
    \centering
    \includegraphics[width=0.57\textwidth, trim=168 130 120 70, clip]{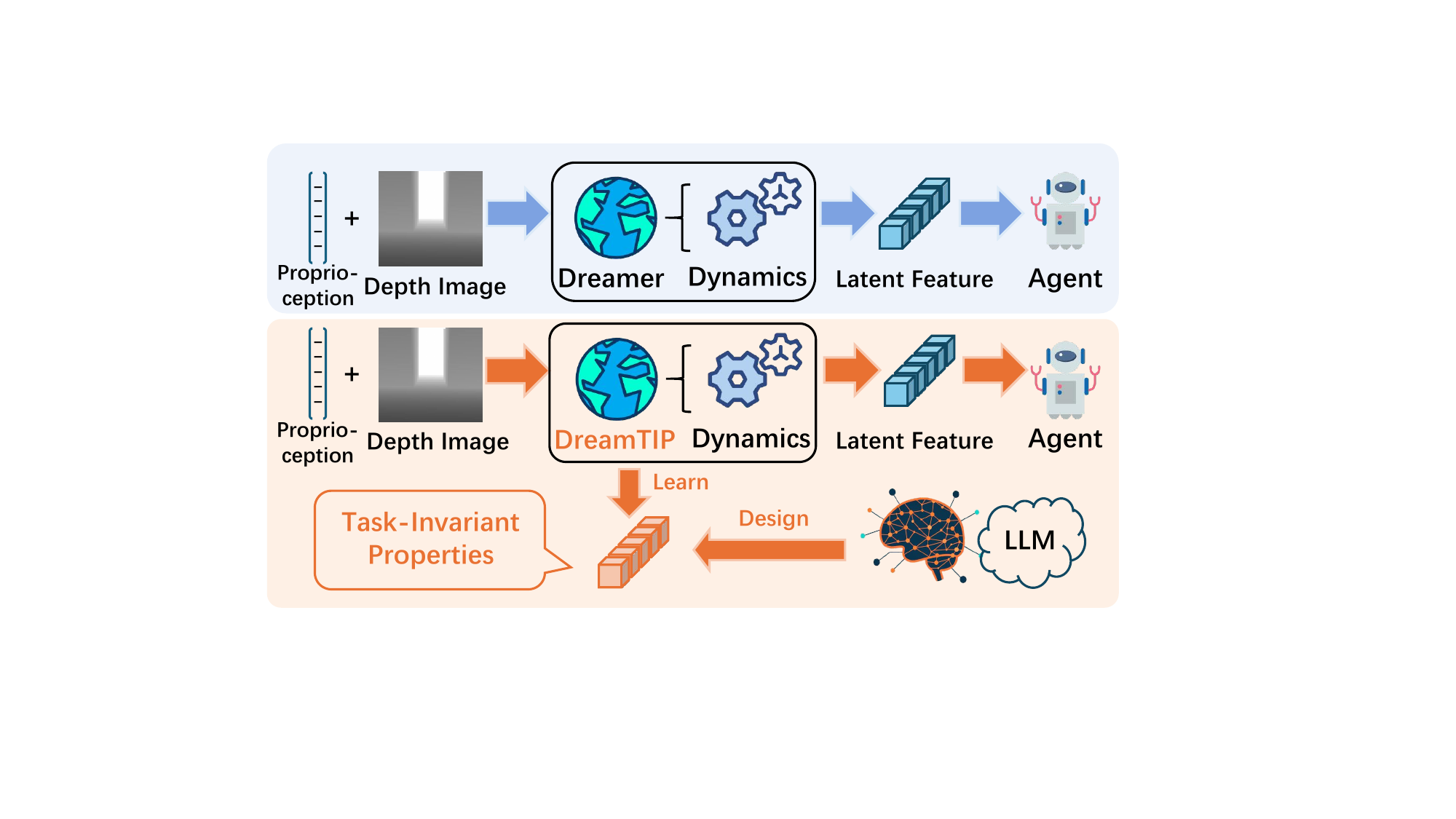}
    \caption{Different Dreamer learning paradigms. The original Dreamer learns environment dynamics by reconstructing observations. DreamTIP, building upon this, also incorporates Task-Invariant Properties designed by an LLM to reduce its over-reliance on underlying dynamics parameters.}
    % \vspace{-10pt}
    \label{fig:Different Dreamer learning paradigms}
\end{figure} 

% Sim-to-Real 迁移研究主要围绕域随机化、域自适应和仿真器增强三个方向展开\cite{zhao2020sim}。域随机化\cite{cheng2024extreme, long2025learning}通过在仿真中引入参数随机性提升策略鲁棒性，但其预定义分布往往难以覆盖真实环境全部变化，泛化能力有限；仿真器增强\cite{wagenmaker2024overcoming}致力于构建高保真仿真环境以减小差异，却依赖精确物理建模且成本高昂，对复杂动态仿真能力存在局限；域自适应\cite{bousmalis2018using, poudel2024recore}通过特征分布对齐缓解域差异，但存在训练不稳定和计算成本高的问题。域自适应方法以学习域不变特征为核心目标，旨在保障策略在跨域环境中的一致性表现\cite{laskin2020curl,gao2025adaworld,mazzaglia2024genrl}。该类方法通过构建泛化性强的表征学习框架，引入额外表征约束，从特征层面缓解仿真与真实之间的动力学差异。例如，Lai 等人 \cite{lai2025world} 提出的 WMP（World Model Perception）利用世界模型提取历史视觉与本体感知的紧凑动态表征，促进策略高效学习；Gu 等人 \cite{gu2024advancing} 提出的 DWL（Denoised World Model），借助特权信息作为辅助监督，基于编码器-解码器架构增强对动力学的建模能力，从而提升机器人在复杂地形中的运动泛化性与适应性。
Research on Sim-to-Real transfer focuses on three main approaches: domain randomization, domain adaptation, and simulator enhancement \cite{zhao2020sim}. Domain randomization \cite{cheng2024extreme, long2025learning} improves robustness by adding parametric randomness during simulation training, but its predefined distributions often fail to cover real-world complexity, limiting generalization. Simulator enhancement \cite{wagenmaker2024overcoming} builds high-fidelity simulations to narrow the reality gap, yet it requires accurate modeling, is costly, and struggles with complex dynamics. Domain adaptation \cite{bousmalis2018using, poudel2024recore} aligns feature distributions across domains to reduce discrepancies, but it often faces training instability and high computational costs.

Domain adaptation methods focus on learning domain-invariant features to maintain consistent policy performance across environments \cite{laskin2020curl, gao2025adaworld, mazzaglia2024genrl}. They incorporate representation constraints to reduce dynamics discrepancies between simulation and reality at the feature level. For example, Lai et al. \cite{lai2025world} proposed World Model Perception (WMP), which uses a world model to extract compact dynamic representations from historical visual and proprioceptive data for efficient policy learning. Gu et al. introduced Denoised World Model (DWL) \cite{gu2024advancing}, employing privileged information as auxiliary supervision in an encoder-decoder structure to improve dynamics modeling and enhance robotic locomotion generalization in complex terrains.

% 尽管如此，现有方法高度依赖于仿真中设定的具体动力学参数。这种依赖性导致策略在面对真实世界的动力学变化时表现脆弱。基于此，我们提出一种基于世界模型策略迁移框架，旨在缓解策略对动力学特征的依赖，实现仅需少量真实样本微调世界模型就能够高效迁移。
Despite this, policies from existing methods heavily rely on the specific dynamic parameters configured in the simulation. This dependency results in fragile performance when facing real-world dynamic variations. To address this, we propose a world model-based policy transfer framework aimed at reducing the policy's reliance on dynamic characteristics, enabling efficient transfer with minimal real-world samples for fine-tuning the world model.

% 具体地，本文基于 Dreamer 框架 \cite{hafner2025mastering} 构建世界模型，用于学习机器人动力学的潜在特征并将其整合至状态空间中，以支撑后续RL策略的学习。为进一步提升策略面对环境变化时的泛化与适应能力，我们提出DreamTIP框架，即在Dreamer训练过程中引入任务不变性属性的世界模型学习方法（Learning Task-Invariant Properties via Dreamer）以引导智能体学习跨任务通用，且对动力学变化保持鲁棒的任务不变性属性，从而减少策略对特定动力学参数的依赖。以腿部机器人运动（locomotion）任务为例，该类特征可体现为接触稳定性（contact-related stability）、地形间隙（terrain clearance）等对动力学鲁棒且与在不同任务上通用的属性。
Specifically, this work builds upon the Dreamer framework \cite{hafner2025mastering} to construct a world model that learns latent features of the robot dynamics and integrates them into the state space, thereby supporting subsequent reinforcement learning (RL) policy training. To further enhance the policy's generalization and adaptation capabilities in the face of environmental variations, we propose the DreamTIP framework: Learning \underline{T}ask-\underline{I}nvariant \underline{P}roperties via \underline{Dream}er (\textbf{DreamTIP}). This approach introduces a world model learning method during Dreamer training that guides the agent to acquire Task-Invariant Properties, which are both transferable across tasks and robust to dynamic changes, thereby reducing the policy's reliance on specific dynamic parameters. Taking legged robot locomotion as an example, such properties can manifest as contact-related stability, terrain clearance, and other dynamics-robust attributes that generalize across diverse tasks.

% 然而，一个关键且尚未被充分探索的问题是：如何准确定义与获取这些任务不变性属性？过去的工作通常依赖于手工设计面向特定任务的中间特征\cite{yang2023recent, liu2025unified}，这种方式需要深厚的领域知识且费时费力，设计成本高昂。为此，本工作引入大语言模型（LLMs）为四足机器人运动任务构建任务不变性属性提取器。LLMs 在海量语料与代码训练中沉淀了丰富的物理常识、行为逻辑及任务规划知识，能够理解高层任务语义并推理出关键属性与成功准则。具体地，我们通过 LLMs 构建 Task-Invariant Extractor，其作用是从特权观测中提取对底层动力学变化不敏感的任务不变性属性，并将其作为世界模型的辅助预测目标。该方法通过构建此类特征表示，使Dreamer的潜在动力学表征对仿真与真实环境间的底层物理差异保持鲁棒性。

However, a critical and underexplored question remains: how to accurately define and acquire Task-Invariant Properties. 
% Previous methods often used manually designed task-specific intermediate features\cite{yang2023recent, liu2025unified}, which require domain expertise and are costly to develop. 
Manual design of task-specific intermediate features\cite{yang2023recent, liu2025unified} is costly, requires deep expertise, and is prone to bias, which limits cross-task generalization. In contrast, Large Language Models (LLMs) leverage their vast pre-trained knowledge to reason about and abstract high-dimensional task descriptions and state observations, uncovering fundamental physical and behavioral principles crucial for task success that human experts might overlook.
To overcome this, we employ LLMs to build a \textbf{Task-Invariant Extractor}. Leveraging their physical and behavioral knowledge, LLMs identify high-level task semantics and extract Task-Invariant Properties from privileged observations. These properties serve as auxiliary prediction targets in the world model, enhancing the robustness of latent representations against sim-to-real physical discrepancies.

%尽管任务不变性属性增强了动力学鲁棒性，仿真与真实环境间仍存在差异。因此，通常需借助少量真实数据对模型进行微调，以校准其参数、匹配真实动力学分布，从而保障策略的最终性能\cite{feng2023finetuning}。然而，微调过程易引发表征崩溃与灾难性遗忘问题\cite{lee2022offline}。为此，本文提出一种高效适应方法，以促进 DreamTIP 在真实环境中的快速适配。
% 具体而言，该方法首先构建一个包含仿真与真实数据的混合回放缓冲池，用于缓解由数据分布差异引起的表征崩溃。随后，我们在适应阶段复制并冻结一个在仿真中预训练的DreamTIP作为参考模型，并通过最小化负余弦相似度来约束后验状态表征的更新。该度量对特征尺度变化不敏感，有助于提升适应稳定性。此外，出于快速适应真实数据分布的考量，在微调过程中冻结DreamTIP的循环模块（recurrent model），以加速世界模型对真实动力学分布的适应
Even though Task-Invariant Properties improve robustness, sim-to-real gaps remain. Fine-tuning with limited real data is often needed to calibrate model parameters toward true dynamics and ensure policy performance \cite{feng2023finetuning}. However, this process risks representation collapse and catastrophic forgetting \cite{lee2022offline}. To overcome these issues, we propose an efficient adaptation method for rapid real-world deployment of DreamTIP. Specifically, our method constructs a mixed replay buffer with both simulated and real data to reduce representation collapse from distribution gaps. During adaptation, we duplicate and freeze a pre-trained DreamTIP model as a reference. Updates to the posterior state are constrained by minimizing negative cosine similarity, a metric insensitive to variations in feature scale, thereby enhancing adaptation stability. Additionally, we freeze DreamTIP’s recurrent module during fine-tuning to accelerate adaptation to real-world dynamics.

% 我们在多种复杂地形（如楼梯、跳跃及攀爬场景）上对所提方法与基线方法进行了对比评估。仿真实验表明，在几乎所有迁移任务中，本文方法均显著优于基线，例如在最高难度（23 cm）的 Crawl 任务中，本文提出方法的平均奖励达到 25.35，远超基线方法的 5.66。进一步在 Unitree Go2 真实机器人平台上的实验也验证了方法的有效性：在 Climb 任务中，基线方法的成功率仅为 10%，而本文方法达到 100% 的成功率。这些结果充分表明，通过提取任务不变性属性并实现高效适应，我们的方法能够显著增强策略在真实环境中的泛化能力，有效弥合仿真与真实之间的差异。
We compared our method with baselines on complex terrains such as stairs, jumps, and scrambles. In simulation, our approach outperformed baselines in nearly all transfer tasks. For instance, in the challenging Crawl task (23 \textit{cm}), our method achieved an average reward of 25.35, far exceeding the baseline’s 5.66. Real-world tests on the Unitree Go2 robot further validated our method: it reached a 100\% success rate in the Climb task (53 \textit{cm}), compared to the baseline’s 10\%. These results demonstrate that extracting Task-Invariant Properties and enabling efficient adaptation significantly improves policy generalization and bridge the sim-to-real gap.

\section{RELATED WORK}
\subsection{Sim to real transfer}
%Sim-to-Real Transfer 旨在解决智能体在仿真环境中训练的策略迁移到真实世界时出现的性能下降问题，其核心挑战在于弥合仿真与真实环境在视觉和动力学方面的领域差异\cite{he2025attention,zhang2024learning,kim2025stage}。主流方法包括域随机化\cite{cheng2024extreme, long2025learning}，通过在仿真中引入视觉或动力学参数的随机变化增强策略泛化性；域适应\cite{bousmalis2018using}，利用对抗训练或特征对齐减少领域间分布差异，从而实现零样本迁移或少量真实数据微调；以及模拟器增强方法，通过构建高保真仿真环境以从源头减小差异。本文工作通过引入世界模型，缓解策略训练对机器人动力学特征的依赖，仅需少量测试环境数据微调世界模型就实现Policy adaptation。
Sim-to-real transfer tackles performance decline when simulation-trained policies face real-world deployment \cite{he2025attention, zhang2024learning, kim2025stage}. The key challenge lies in bridging visual and dynamic gaps between domains. Main approaches include: domain randomization \cite{cheng2024extreme, long2025learning}, improving generalization through varied simulation parameters; domain adaptation\cite{bousmalis2018using}, aligning feature distributions for zero-shot transfer or efficient fine-tuning; and simulator enhancement\cite{wagenmaker2024overcoming}, building high-fidelity environments. This paper introduces a world model to reduce dynamics-specific dependency, enabling effective adaptation with minimal real-world data.

\subsection{World model for robotics}
% 世界模型通过建模环境动力学支持预测与规划，显著减少了对真实交互数据的依赖\cite{hansen2023td}。例如，DreamerV3\cite{hafner2025mastering} 通过潜在动力学预测与多尺度优化，在多项任务中实现了高效稳定的训练；DayDreamer\cite{wu2023daydreamer} 仅凭有限真实交互就使机器人获得复杂行为并实现在线适应；Lai 等人\cite{lai2025world} 所提的 WMP 方法则从多模态感知中提取紧凑表征以提升策略学习效率。
% 基于此，本文进一步探索世界模型在跨域泛化中的作用，提出通过显式学习任务不变性属性，增强模型对动力学变化的鲁棒性，以更好地促进仿真到现实的迁移。
World models significantly reduce the reliance on real-world interaction data by modeling environmental dynamics to support prediction and planning \cite{hansen2023td}. For instance, DreamerV3 \cite{hafner2025mastering} achieves efficient and stable training across multiple tasks through latent dynamics prediction and multi-scale optimization; DayDreamer \cite{wu2023daydreamer} enables robots to acquire complex behaviors and achieve online adaptation with limited real-world interactions; and the WMP method proposed by Lai et al. \cite{lai2025world} extracts compact representations from multimodal perception to enhance policy learning efficiency.
Building on these advances, we further explore the role of world models in cross-domain generalization and propose explicitly learning Task-Invariant Properties to enhance the model's robustness to dynamic variations, thereby better facilitating simulation-to-real transfer.

\subsection{LLM-driven robot skill learning}
%大语言模型（LLMs）在四足机器人中的应用主要体现在奖励建模、运动控制和指导表征学习三个方面。
% 在奖励建模方面，Eureka\cite{ma2023eureka}系列的工作展示了 LLMs 在自动生成机器人控制人物奖励函数上的潜力。
% 在运动控制方面，研究者既探索了利用 LLMs 将自然语言转化为中间运动表示（如足端接触模式或中层指令），再由强化学习控制器执行\cite{tang2023saytap,cheng2024navila}，也尝试直接将 LLM 作为低层反馈控制器来生成关节轨迹，从而推动四足机器人实现语言到动作的直接映射\cite{wang2024prompt}。
% 在大语言模型（LLMs）与表示学习结合，用于改进状态表征和控制策略中。Wang \cite{wang2024llm}等提出 LESR (LLM-Empowered State Representation)，利用 LLMs 的知识生成状态表征与内在奖励，从而提升值函数映射的连续性和采样效率。相比之下，我们的方法利用LLMs的丰富知识与推理能力，专注于提取与任务成功密切相关的不变性特征，从而提升机器人策略的泛化能力。
Large language models are applied in quadruped robots across three main areas: reward modeling, motion control, and representation learning. In reward design, works like Eureka \cite{ma2023eureka} show LLMs can automatically generate reward functions. For motion control, some studies use LLMs to convert language into intermediate commands (e.g., foot contact patterns) executed by reinforcement learning controllers \cite{tang2023saytap, cheng2024navila}, or even directly output joint trajectories \cite{wang2024prompt}. In representation learning, methods such as LESR \cite{wang2024llm} employ LLMs to improve state representations and intrinsic rewards, enhancing policy generalization and efficiency. In contrast, our approach utilizes LLMs' rich knowledge and reasoning capabilities to extract Task-Invariant Properties closely tied to task success, ultimately enhancing the generalization ability of robot policies.

\section{BACKGROUND}
\begin{figure*}[t]
    \centering
    \includegraphics[width=0.99\textwidth, trim=105 175 185 165, clip]{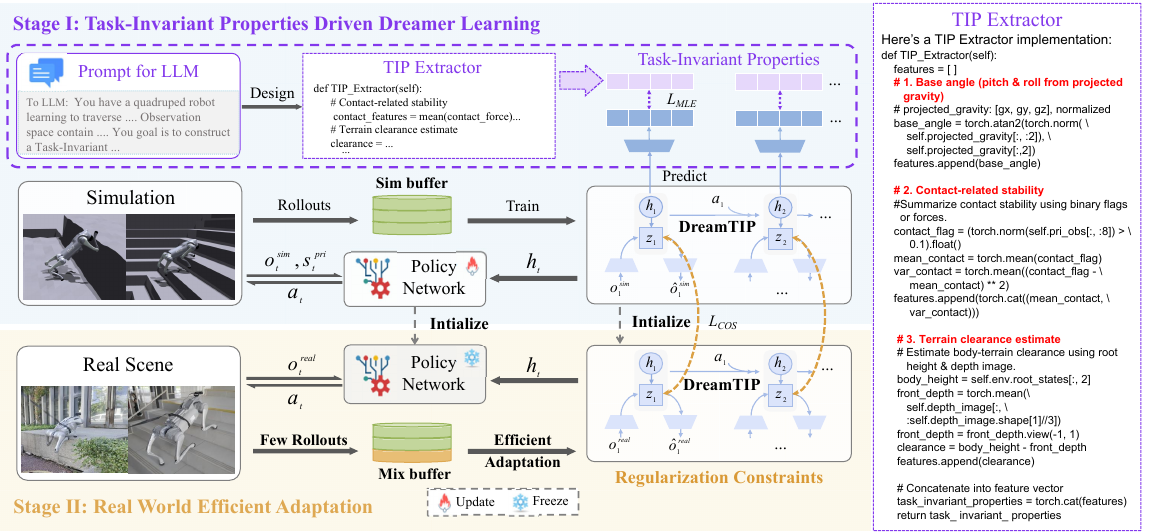} 
    \caption{Overview of the proposed framework. The framework consists of two stages: In the first stage, DreamTIP is employed in a simulation environment to learn Task-Invariant Properties; In the second stage, it  adapts to the dynamics distribution in the physical environment with only a few rollouts.}
    % \vspace{-10pt}
    \label{fig:Framework}
\end{figure*} 
\subsection{Problem Formulation}
%四足机器人运动技能学习是一个部分可观测马尔可夫决策过程 represented by the tuple(S, O, A, P, R, γ),S表示状态，O表示观测，A是动作空间，P是状态转移方程，R是奖励函数，γ是折扣因子，目标是最大化expected return：$\mathbb{E}\left[\sum_{t=0}^{\infty} \gamma^{t} r_{t}\right]$
%具体而言，在仿真环境中，除了可以获得包括本体感知信息与深度图像的观测ot∈O，我们还可以直接获得特权信息st∈S。
The learning of locomotion skills for legged robots can be formulated as a Partially Observable Markov Decision Process (POMDP), represented by the tuple  \( (\mathcal{S}, \mathcal{O},\mathcal{A}, \mathcal{P}, \mathcal{R}, \gamma) \), where \( s_t \in \mathcal{S} \) is the state space, which encapsulates the full dynamical state of the robot and its environment. \( o_t \in \mathcal{O} \) is the observation, \( a_t \in \mathcal{A} \) is the action space, \( \mathcal{P} \) is the state transition function, \( \mathcal{R} \) is the reward function, and \( \gamma \in (0, 1)\) is the discount factor. The goal of the agent is to learn a policy that maximizes its cumulative discounted return $R_t = \sum_{i=0}^{\infty} \gamma^{i} r_{t+i}$.

\subsection{Dreamer-Augmented Policy Optimization}
%Dreamer\cite{hafner2025mastering}通过学习一个潜在动力学模型，从像素或状态观测中提取抽象的环境动态表示。本工作中使用的是dreamer的变体，该模型的核心组件为其递归状态空间模型（Recurrent State-Space Model, RSSM），主要包括以下四个部分：
Dreamer \cite{hafner2025mastering} learns a latent dynamics model to extract abstract representations of environmental dynamics from pixel or state observations. In this work, we employ a variant of Dreamer whose core component is the Recurrent State-Space Model (RSSM), which primarily consists of the following four parts: Recurrent model $h_{t} = f_{\theta}(h_{t-1}, z_{t-1}, a_{t-1})$, Encoder $z_{t} \sim q_{\theta}(z_{t} \mid h_{t}, o_{t})$, Dynamic predictor $\hat{z}_{t} \sim p_{\theta}(\hat{z}_{t} \mid h_{t})$ and Decoder $\hat{o}_{t} \sim p_{\theta}(\hat{o}_{t} \mid h_{t}, z_{t})$.

% 通过最小化负变分下界（ELBO）来联合学习整个模型：
Dreamer jointly learns the entire model by minimizing the negative Evidence Lower Bound (ELBO):
\begin{equation}
\label{eq:dreamer loss}
\begin{aligned}
\mathcal{L}_{D}(\theta) \doteq & \mathbb{E}_{q_{\theta}} \Big[ \sum_{t=1}^{T} \big( -\ln p_{\theta}(o_t \mid h_t, z_t) \\
 & + \beta \, \text{KL} \big[ q_{\theta}(z_t \mid h_t, o_t) \parallel p_{\theta}(\hat{z}_t \mid h_t) \big] \big) \Big],
\end{aligned}
\end{equation}
where $\beta$ is a hyperparameter.

%PPO\cite{schulman2017proximal}是一种Actor-Critic架构的策略优化算法，其目标是通过优化策略梯度来学习一个能最大化累积回报的策略$\pi_{\theta}(a_t \mid o_t)$。Inspired by WMP \cite{lai2025world},我们将Dreamr输出的确定性状态表征$h_t$与当前观测$o_t$共同作为 Actor-Critic 网络的输入。因此PPO的优化目标重定义为学习一个能最大化累积回报的策略$\pi_{\theta}(a_t \mid h_t, o_t)$
PPO \cite{schulman2017proximal} is a policy optimization algorithm based on the Actor-Critic framework, which aims to learn a policy $\pi_{\theta}(a_t \mid o_t)$ that maximizes cumulative returns by optimizing policy gradients. Inspired by WMP [17], we first input the observation $o_{t-1}$, which consists of proprioceptive data and depth images, into Dreamer. 
% The recurrent model module then infers the deterministic state representation $h_t$. 
% \color{red}
And then the hidden state $h_t$ is computed by the recurrent model (RNN) based on the previous deterministic state $h_{t-1}$, stochastic state $s_{t-1}$, and action $a_{t-1}$. It encodes the deterministic history of the environment’s dynamics, capturing the complete temporal evolution from the initial state to the current time step.
% \color{black}
Subsequently, incorporate the $h_t$ along with the current observation $o_t$ (without depth image) as inputs to the Actor-Critic network. Consequently, the optimization objective of PPO is redefined as learning a policy $\pi_{\theta}(a_t \mid h_t, o_t)$ that maximizes cumulative returns.

\section{METHOD}
\subsection{Overview}

%本框架首先利用大型语言模型分析任务描述与状态观测空间，构建任务不变性属性提取器，用于将特权状态转换为对应的任务不变性属性。随后，我们在 Dreamer 基础上引入一个额外预测头，构建其改进版本 DreamTIP，以显式地学习这些任务不变性属性。在部署阶段，框架复制并冻结预训练的 DreamTIP 参数作为参考模型。在微调过程中，采用混合真实数据与仿真数据的回放缓冲池进行离线更新，并通过最大化微调模型与冻结参考模型在同一观测下所得潜在表征之间的余弦相似度，为模型适应过程提供有效的正则化约束，从而仅需少量真实数据即可实现策略的高效适应.
In this section, we present the proposed framework, termed Learning Task-Invariant Properties via Dreamer. As shwon in Fig.~\ref{fig:Framework}, this framework first leverages a large language model to analyze task descriptions and state observation spaces, constructing a \textbf{TIP Extractor} to convert privileged states into corresponding \textbf{Task-Invariant Properties}. Subsequently, we introduce an additional predictor on the Dreamer architecture, developing an improved version termed \textbf{DreamTIP}, to explicitly learn these Task-Invariant Properties. During the deployment phase, the framework duplicates and freezes the pre-trained DreamTIP parameters to serve as a reference model. During the adaptation process, a mixed replay buffer comprising both real and simulated data is utilized for offline updates, while regularization constraints are applied to the adaptation process. This approach enables efficient adaptation with minimal real-world data requirements.

\subsection{Task-Invariant Properties} 
%为了解决人工设计特征的局限性，本工作利用大语言模型的先验知识与推理能力，构建一个任务不变性属性提取器，将原始特权观测信息转换成与任务成功密切相关的任务不变性属性。具体而言，如图2所示，我们在训练开始前，预先将高层的任务描述 $Text$ 与包含特权信息的完整状态观测空间&O&作为输入，引导LLM推理并输出一个任务不变性属性转换函数$F_{extrator}$。该函数&F_{extrator}&能够从原始特权观测$s_{t}$中提取出与跨任务通用且对动力学变化不敏感的任务不变性属性$f_{t}$，其形式化定义为：
Manually designing these properties entails high costs and inherent limitations. To overcome these constraints, this work leverages the prior knowledge and reasoning capabilities of LLMs to construct a \textbf{TIP Extractor}. This module converts raw privileged observational information into \textbf{Task-Invariant Properties} that are closely correlated with task success. Specifically, as illustrated in Fig.~\ref{fig:Framework}, prior to training, we provide the high-level task description $I_{text}$ and the complete state observation space $I_{priv}$ containing privileged information as inputs to guide the LLM in reasoning and outputting a properties transformation function $TIP_{extrator}$. This function $TIP_{extrator}$ is capable of extracting Task-Invariant Properties $f_{t}$ from the raw privileged observations $s_{t}$, which are both generalizable across tasks and insensitive to variations in dynamics. The process is formally defined as:
\begin{equation}
\label{eq:task-invariant property extractor}
TIP_{extrator}=\text{LLM}(I_{text},I_{priv}),
\end{equation}
\begin{equation}
\label{eq:task-invariant property}
f_t=TIP_{extrator}(s_t).
\end{equation}

%在此基础上，为将提取的任务不变性属性$f_t$融入世界模型的学习过程并增强其迁移能力，我们在Dreamer基础上\cite{hafner2025mastering}提出其改进版本DreamTIP，该框架的核心是引入一个任务不变性属性预测头，旨在使世界模型能够从自身的潜在状态中推断出与LLM所定义的同一任务不变性属性$f_t$。该Predictor由MLP实现，以时刻$t$DreamTIP的循环状态 $ht$与随机状态表征$zt$的拼接作为输入，预测当前时刻的任务不变性属性估计$\hat{f_t}$,即：$\quad \hat{f}_t \sim p_{\phi}({f}_t \mid h_t, z_t)$。由此，$f_t$将作为DreamTIP的一个额外预测目标，通过最大似然损失函数进行重构。%根据式x，DreamTIP的优化目标可重新表达为：
Building upon this foundation, and to integrate the extracted Task-Invariant Properties $f_t$ into the world model's learning process while enhancing its transfer capabilities, we propose DreamTIP, an improved version of the Dreamer framework \cite{hafner2025mastering}. The core innovation of this architecture lies in the introduction of a properties predictor, designed to enable the dreamer to infer the same Task-Invariant Properties $f_t$ from its own latent states. This predictor, implemented as a Multilayer Perceptron (MLP), takes as input the concatenation of DreamTIP’s recurrent state $ht$ and stochastic state representation $z_t$ at time step $t$, and outputs an estimate $\hat{f_t}$ of the current Task-Invariant Properties, then Eq. \ref{eq:dreamer loss} can be rewritten as:
\begin{equation}
\label{eq:DreamTIP traing}
\begin{aligned}
\mathcal{L}_{train}(\theta) \doteq \mathcal{L}_{D} (\theta) -\mathbb{E}_{q_{\theta}}  \Big[\sum_{t=1}^{T} \ln p_{\theta}(f_t \mid h_t, z_t) \Big],
\end{aligned}
\end{equation}
where the second term is $L_{MLE}$.

%该方法通过引入任务不变性属性，促使世界模型学习跨任务通用且对动力学扰动鲁棒的表征，从而有效提升世界模型在未知真实环境中的泛化能力与表现一致性。
% \color{red}
By incorporating Task-Invariant Properties, this approach encourages the world model to learn representations that are both generalizable across tasks and robust to dynamic disturbances, thereby enhancing its transfer capability and behavioral consistency in previously unknown real-world environments.
% \color{black}

\subsection{Real-World Efficient Adaptation} 
%在DreamTIP的仿真训练过程中，我们同步收集仿真轨迹并构建仿真经验回放缓存（sim buffer）。待模型训练收敛后，复制并冻结其参数，得到冻结的DreamTIP $M_sg$，并将其与冻结的策略模型$\pi$共同部署于四足机器人实体中。随后，通过在真实环境中进行交互，持续采集真实轨迹数据并不断并入原有的 sim buffer，构成混合经验缓存（Mix buffer），用于后续DreamTIP的”离线适应更新“。该混合缓冲机制旨在通过平衡新旧数据分布，约束模型更新幅度，避免因过度优化少量真实样本而破坏或遗忘已学习到的动力学转移表征，从而缓解灾难性遗忘、表征坍塌与过度拟合等问题，平稳提升模型对真实动力学的适应能力。
%在适应训练阶段，世界模型虽具有较高的数据利用效率，但其微调效果对真实数据的质量和数量依然敏感，直接进行全参数微调在数据有限时容易表现不佳。然而，真实环境中的数据采集成本高且样本有限。为以少量真实轨迹实现世界模型对真实动力学的快速适应，在实际的适应更新过程中，我们冻结 DreamTIP $M$中的 Recurrent Model 模块。此举主要出于快速适应真实数据分布的考量，以加速世界模型对真实动力学分布的适应。
During the simulation training phase of DreamTIP, we concurrently collect simulated trajectories to construct a simulated experience replay buffer (\textbf{Sim buffer}). Upon model convergence, its parameters are duplicated and frozen to obtain a fixed DreamTIP model $M_{sg}$, which is deployed on the quadruped robot alongside a frozen policy model $\pi$. Subsequently, through interaction in the real environment, real trajectory data is continuously collected and incrementally merged into the original sim buffer to form a mixed replay buffer (\textbf{Mix buffer}). This mix buffer is utilized for subsequent offline adaptation updates in DreamTIP. The hybrid buffer mechanism is designed to balance the distribution of old and new data, thereby constraining the magnitude of model updates. This approach avoids over-optimization on limited real samples, which could otherwise disrupt or cause the forgetting of previously learned dynamic transition representations. Thereby, it mitigates issues such as catastrophic forgetting, representation collapse, and overfitting, and consistently enhances the model's adaptability to real-world dynamics.

Although the world model demonstrates high data efficiency during adaptation, its fine-tuning performance is still highly dependent on the quality and scale of real-world data, while real-world data collection is costly and sample availability is typically restricted. Furthermore, full parameter fine-tuning under limited data conditions often results in suboptimal outcomes. To enable rapid adaptation of DreamTIP to real dynamics using a small number of real trajectories, we freeze the Recurrent Model module within DreamTIP during the adaptation update process. This strategy is primarily motivated by the need to accelerate the world model’s alignment with the distribution of real data, thus promoting faster adaptation to real-world dynamics.

%为进一步提升适应训练过程的稳定性，受TWIST \cite{yamada2024twist} 的师生蒸馏世界模型和 LS-UNN \cite{BEAUSSANT2025104862}中源模型监督机制工作的启发，我们提出Consistency Constraint-Based Efficient Adaptation方法，通过引入正则化约束提升模型微调过程的稳定性。具体而言，在仿真中得到预训练的DreamTIP模型后$M$，我们复制并冻结$M$的参数得到参考世界模型$M_sg$，随后将$M_sg$在每一个时间步的随机状态表示作为监督信号，以约束$M$在真实数据上的适应过程。具体而言，对于每一时间步的观测输入$o_t$，将其分别输入至$M_sg$的编码器与待微调适应的$M$的编码器中，获得相应的随机状态表示 $Z_t^sg$与$z_t$。通过最小化负余弦相似度损失，约束两个表征在方向上的对齐。该损失函数对特征尺度不敏感，侧重于语义一致性而非数值匹配\cite{barz2020deep}，为核心的重建损失提供了有效的正则化约束。这一约束旨在稳定模型在少量真实数据上的更新过程，防止因过度优化重建损失而导致的表征崩溃或偏离预训练得到的有效表征空间。在适应过程中，策略网络$\pi$的参数始终保持冻结。该方法通过这种正则化机制，在少量真实数据下实现了稳定、高效的世界模型适应。综上所述，适应更新过程的完整损失函数可定义为：

To further enhance the stability of adaptation training, inspired by the teacher–student distillation framework in TWIST \cite{yamada2024twist} and the source model supervision strategy in LS-UNN \cite{BEAUSSANT2025104862}, we propose a Regularization Constraints-based stable adaptation method. The key idea is to stabilize model adaptation by introducing regularization constraints. Specifically, we first obtain a pre-trained DreamTIP model $M$ from simulation, then duplicate and freeze its parameters to construct a reference world model $M_{sg}$. At each timestep $t$, the stochastic state representation of $M_{sg}$ serves as a supervisory signal to guide the adaptation of $M$ on real data.

Concretely, during the adaptation process, for each observation $o_t$ at timestep $t$, both the frozen reference model $M_{sg}$ and the adaptable model $M$ encode it into stochastic state representations, denoted as $z_t^{sg}$ and $z_t$ respectively. A negative cosine similarity loss is minimized to align their directions, enforcing semantic consistency while remaining insensitive to feature scales \cite{barz2020deep}. 
This regularization term plays a critical role in complementing the reconstruction loss. While the reconstruction objective encourages the world model to fit real-world data distributions, limited data availability may drive the learned representation towards collapseor cause it to deviate from the well-structured latent space established during pre-training. The regularization constraint addresses this issue by stabilizing the adaptation process and preserving the quality of the latent representations.

According to Eq. \ref{eq:dreamer loss}, the complete loss function for the adaptation process can be refined as:
\begin{equation}
\label{eq:quick adaptation}
\begin{aligned}
\mathcal{L}_{Adapt}(\theta) \doteq \mathcal{L}_{D} (\theta) -\mathbb{E}_{q_{\theta}}  \Big[\sum_{t=1}^{T} \frac{z_t \cdot z_t^{sg}}{\|z_t\| \cdot \|z_t^{sg}\|} \Big].
\end{aligned}
\end{equation}
where the second term is $L_{COS}$.

During adaptation, the policy network $\pi$ remains frozen. With this mechanism, our approach achieves stable and efficient world model adaptation using only a small amount of real-world data.

\begin{figure*}[t]
    \centering
    \includegraphics[width=0.92\textwidth, trim=75 10 125 15, clip]{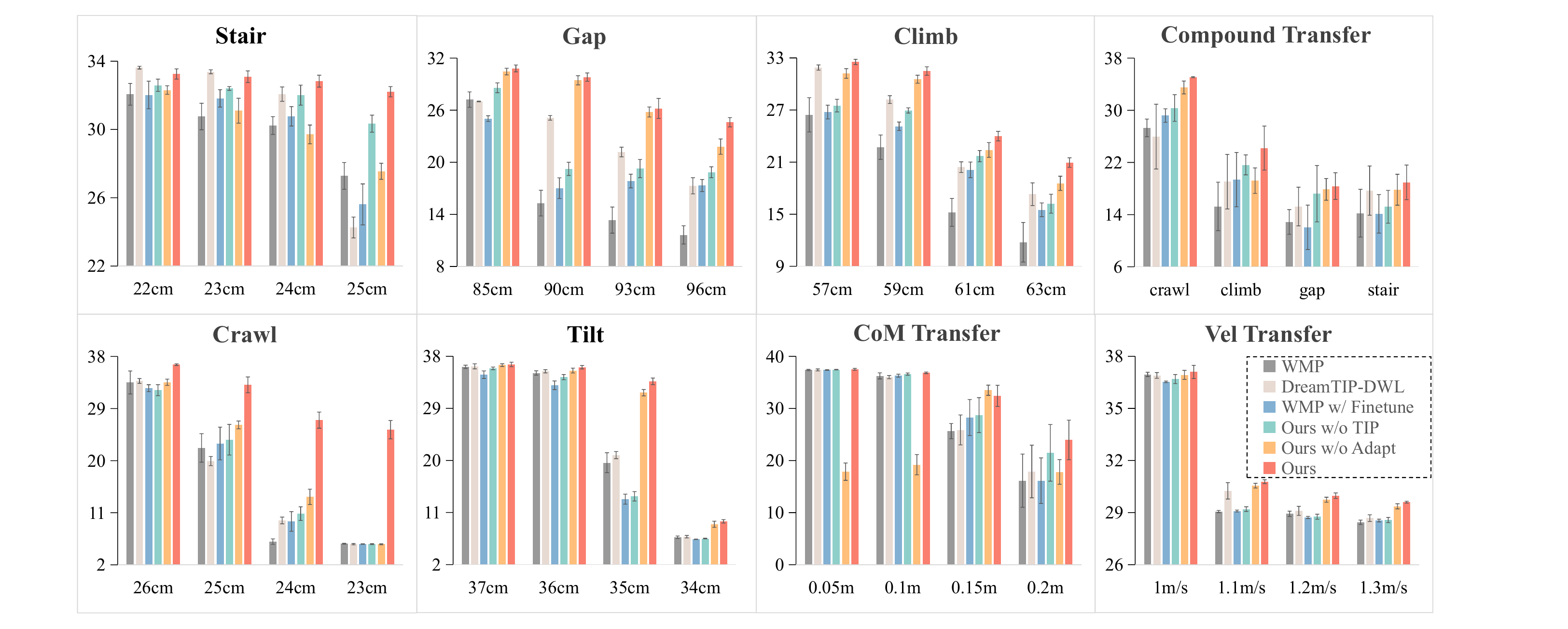}
    \caption{Performance comparison of various methods on eight transfer tasks in simulation. The evaluation metric is the average cumulative reward over 100 trajectories per task. Our method outperformed all other baselines across the board. The vertical axis represents the average trajectory reward, while the horizontal axis indicates the varying levels of task difficulty. The results are obtained through testing over 100 trajectories with 3 different random seeds.}
    \label{fig:Main results}
\end{figure*} 

\section{EXPERIMENTS}
% 为了验证所提出方法在迁移任务中的有效性，我们分别在仿真环境与真实场景中开展了实验。仿真实验基于 NVIDIA Isaac Gym 平台搭建，构建了包含 stair、gap、climb、crawl、tilt 以及 rough flat 在内的六类典型地形。实验采用 Unitree Go2 四足机器人，其动作空间为12维，对应12个关节的目标位置；观测空间包括本体感知信息（如基础角速度、重力投影方向、关节位置与速度）以及深度图像。特权信息除涵盖上述观测内容外，还补充了线速度、高程图、摩擦系数、质心位置以及足端接触力等物理状态。在真实世界验证中，我们同样使用 Unitree Go2 机器人进行测试。所有方法都是直接在板载的Orin Nano上运行，深度图由D435i获取，We preprocess the noisy depth images with spatial and temporal filters to narrow the visual sim-to-real gap \{zhuang2023robot}
\begin{figure}[t]
    \centering
    \includegraphics[width=0.999\columnwidth, trim=130 120 240 120, clip]{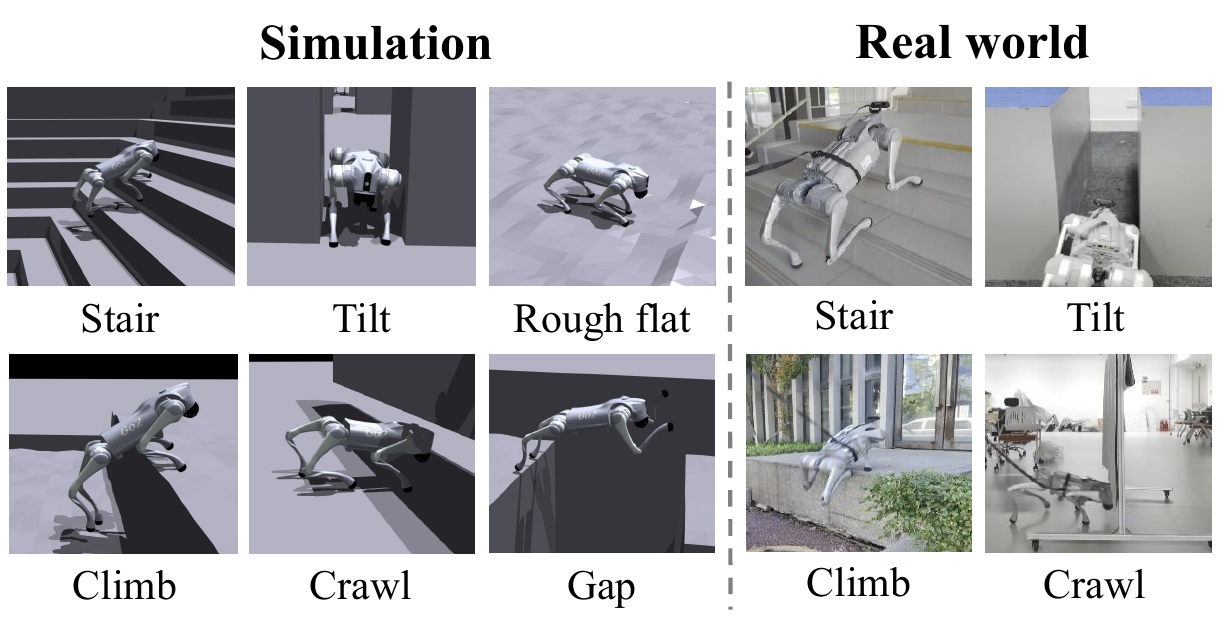}
    \caption{Illustrations of terrain settings in simulation and real-world evaluation.}
    \label{fig:task fig}
\end{figure} 
\subsection{Experiment Setting} %图4展示了我们在仿真及真机实验中的地形任务设置
To validate the effectiveness of the proposed method in transfer tasks, experiments were conducted in both simulated and real-world environments. Fig.~\ref{fig:task fig} illustrates the configuration of terrain tasks employed in our study, covering both simulation and real-world evaluation. The simulation experiments were built on the Isaac Gym environment. We employed Unitree Go2 robots for the experiments, whose action space is 12-dimensional, corresponding to the target positions of the 12 joints. The observation space includes proprioceptive information such as base angular velocity, direction of gravity projection, joint positions and velocities, along with depth images. Beyond the observation variables mentioned above, the privileged information also incorporates physical states such as linear velocity, elevation maps, friction coefficients, center of mass position, and foot contact forces. In real-world evaluation, all methods were deployed and executed directly on the onboard Orin Nano of the Go2 robot, utilizing depth images captured by the D435i camera, which were preprocessed with spatial and temporal filters to mitigate the visual sim-to-real gap \cite{zhuang2023robot}.

%例如，在四足机器人穿越地形任务中，LLM生成的任务无关属性（TIP）如图X上部所示，其揭示出“climb，stairs，gap”等任务均依赖两项共同约束：一是维持足够的离地间隙以规避物理碰撞，二是保持足端接触稳定性以防止滑倒失稳
% 本工作设计的
% \color{red}
\textbf{TIP Extractor:} In quadruped robot locomotion tasks mentioned in this paper, the TIP generated by the LLM, as shown in the right part of Fig.~\ref{fig:Framework}, reveal that tasks such as climb, stairs, and gaps share below critical common constraints: maintaining sufficient terrain clearance to avoid physical collisions, and preserving foot contact stability to prevent slipping and instability.
% \color{black}

\textbf{Reward functions:} We adopt a reward function similar to that of Cheng et al. \cite{cheng2024extreme}, which encourages the robot to follow the commanded velocities while penalizing velocities along other axes, excessive joint torques, accelerations, and collisions. In addition, we introduce two extra reward terms: penalizing joint deviations from the normal standing posture, and encouraging smoothness of joint torques \cite{kumar2021rma}. We find that these designs are beneficial for sim-to-real transfer.

\textbf{Simulation settings:} A total of eight transfer tasks were constructed in the simulation environment for comprehensive evaluation. During training, the center of mass was randomized within [-0.05 \textit{m}, 0.05 \textit{m}], and the velocity command varied over [0, 1] \textit{m/s}. Across the five terrain transfer tasks (stair, gap, climb, crawl, and tilt), each task is evaluated under difficulty levels not encountered during training. To further evaluate the robot’s adaptation capability under unseen variations in mass distribution and velocity commands, the CoM Transfer and Velocity Transfer tasks were conducted on rough flat terrain. Specifically, in the CoM Transfer task, a mass-center offset $\Delta a$ was introduced, resulting in an adjusted range: [-0.05-$\Delta a$, 0.05-$\Delta a$]$\cup$[-0.05+$\Delta a$, 0.05+$\Delta a$]. In the Compound Task, we set a mass center offset ($\Delta a$ = 0.1 \textit{m}) and a fixed velocity command (1.1 \textit{m/s}) for all four tasks: Gap (75 \textit{cm}), Climb (40 \textit{cm}), Crawl (25 \textit{cm}), and Stair (18 \textit{cm}). This configuration induced a more challenging scenario.

\begin{figure*}[t]
    \centering
    \includegraphics[width=0.96\textwidth, trim=75 120 100 90, clip]{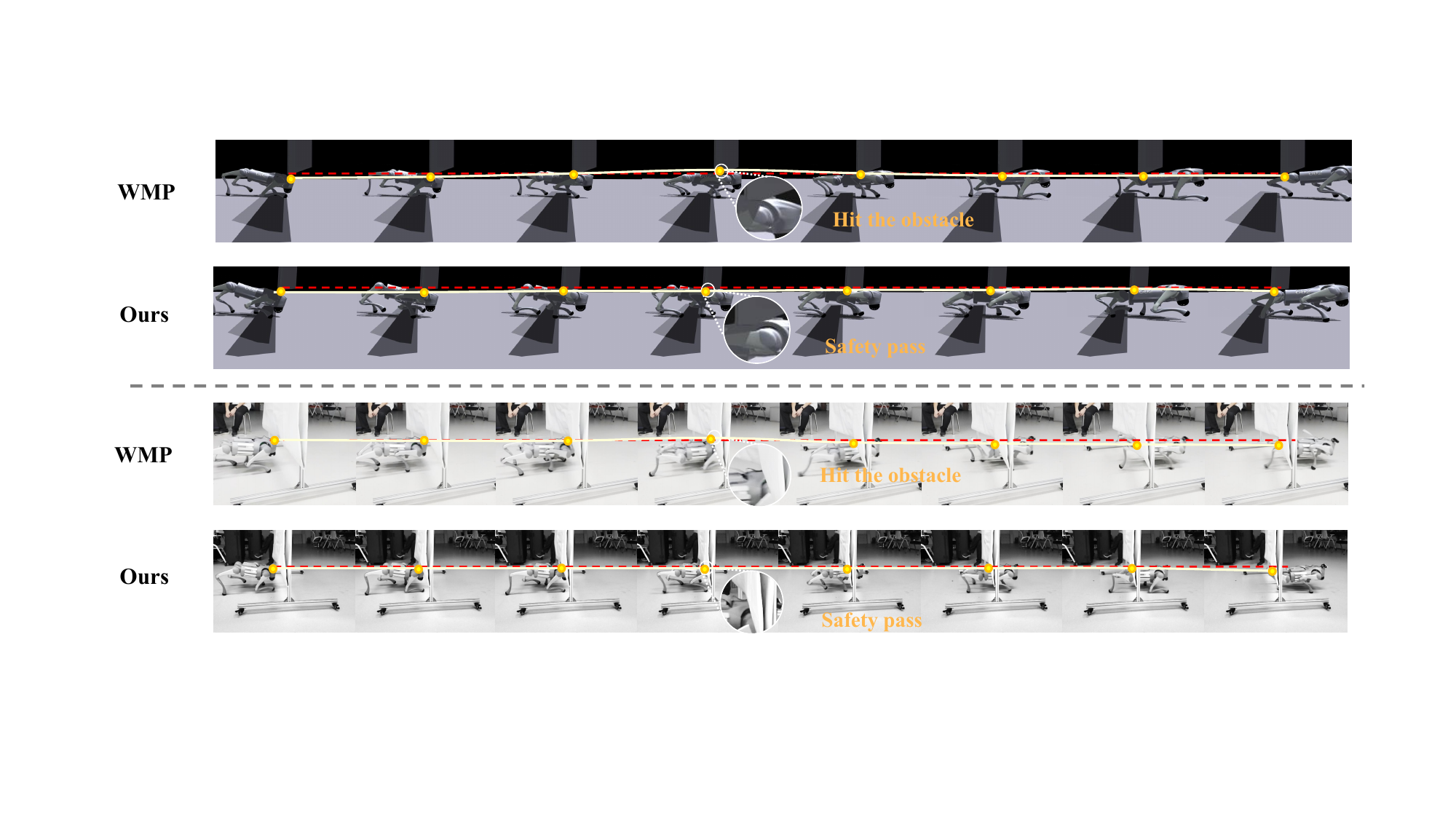}
    \caption{Performance comparison of various methods on the crawl task across simulated and real environments. Simulation environment (top, above gray dashed line) and real-world environment (bottom). Red line: obstacle height; Yellow dots: robot dog's traversal height at the obstacle. With the obstacle height set to 25 \textit{cm} in both environments, the Baseline method encounters collisions with its head when passing through the obstacles, whereas our method traverses safely. This demonstrates the superior task transfer performance of our method, as well as the consistency in its sim-to-real effectiveness.}  
    \label{fig:comparison}
\end{figure*} 

\subsection{Simulation Evaluation}
%仿真任务设置：
%我们在仿真环境中设置了八种迁移任务以进行全面评估。训练过程中，机器人的质心随机化范围为 [-0.05 m, 0.05 m]，速度指令的范围为 [0, 1] m/s。具体测试任务包括 stair、gap、climb、crawl 和 tilt 五种类型，每种任务均在测试阶段引入了训练中未出现过的难度等级。此外，为评估机器人在面对未知质心变化或速度指令变化时的迁移能力，我们在 Com Transfer 和 Velocity Transfer 任务中使用了 rough flat 地形作为测试环境。在 Com Transfer 任务中，我们引入质心偏移量 Δa，此时质心变化范围调整为：[-0.05 – Δa, 0.05 – Δa] ∪ [-0.05 + Δa, 0.05 + Δa]。对于 Compound Task，我们在 gap（75 cm）、climb（40 cm）、crawl（25 cm）和 stair（18 cm）四个任务中同时引入训练中未曾出现的质心偏移量 Δa = 0.1，并将速度指令统一设置为 1.1 m/s，从而构建更具挑战性的复合迁移场景。
The methods involved in the experiments conducted in this paper are as follows. WMP \cite{lai2025world} and DreamTIP-DWL \cite{gu2024advancing} are the two baseline methods. The specific descriptions are as follows:

% \textbf{WMP:} 遵循 Lai 等人 \cite{lai2025world} 提出的训练范式，在仿真环境中训练 Dreamer 模型，且在跨任务迁移过程中不进行任何适应更新。
% \textbf{DreamTIP-DWL:} 基于 Gu 等人 \cite{gu2024advancing} 的训练框架，在DreamTIP训练过程中，利用$h_t$与$z_t$预测特权信息$s_t$，而非任务不变性属性。迁移过程中不执行模型更新。
% \textbf{WMP w/ finetune:} 在 WMP 基础上，使用混合经验回放池（mix buffer）进行适应更新，训练过程中冻结序列模型（Sequence Model），且不施加正则化约束。
% \textbf{Ours w/o TIP:} 本方法在 DreamTIP 训练阶段不学习任务不变性属性，而在迁移任务中借助 mix buffer 对模型进行适应更新。
% \textbf{Ours w/o Adapt:} 本方法在 DreamTIP 训练阶段学习任务不变性属性，但在迁移过程中不进行任何适应更新。
% \textbf{Ours:} 本文所提方法,在 DreamTIP 训练阶段学习任务不变性属性，并在迁移过程中使用 mix buffer 对模型进行适应更新。
\textbf{WMP:} Following the training paradigm proposed by Lai et al. \cite{lai2025world}, this method differs from DreamTIP by omitting both Task-Invariant Properties and adaptation updates.

\textbf{DreamTIP-DWL:} According to the training framework introduced by Gu et al. \cite{gu2024advancing}, DreamTIP predicts privileged information during DreamTIP training instead of Task-Invariant Properties, and performs no adaptation updates.

\textbf{WMP w/ Finetune:} Based on WMP, this variant utilizes adaptation updates. The sequence model is frozen during adaptation, and no regularization constraints are applied.

\textbf{Ours w/o TIP:} The proposed method without learning Task-Invariant Properties during the training phase.

\textbf{Ours w/o Adapt:} The proposed method without DreamTIP Adaptation during the transfer process.

\textbf{Ours:} The proposed method.

\begin{table}[t]
\centering
\setlength{\tabcolsep}{2pt}
\begin{tabular}{ccccc}
\toprule
Methods & Stair (16\textit{cm})& Climb (52\textit{cm})& Tilt (33\textit{cm})& Crawl (25\textit{cm})\\
\midrule
WMP &100\% &10\% &40\% & 70\% \\ 
Ours w/o Adapt &100\% &90\% &50\% &80\%\\ 
Ours &\textbf{100\%} &\textbf{100\%} &\textbf{80\%} &\textbf{100\%}\\ 
\bottomrule
\end{tabular}
\caption{Real-world evaluation. The success rate was employed as the evaluation metric in this study. The results were statistically derived from 10 independent trials conducted for each task.}
\label{tab:Real World Evaluation}
% \vspace{-15pt}
\end{table}

\begin{table*}[t]
\centering
\addtolength{\tabcolsep}{-2pt}
\begin{tabular}{ccccccccc}
\toprule
Methods& \multicolumn{4}{c}{Climb} & \multicolumn{4}{c}{Tilt}\\
\cmidrule(lr{0.3em}){2-5} \cmidrule(lr{0.3em}){6-9} 
 -& 57cm& 59cm& 61cm& 63cm& 37cm& 36cm& 35cm &34cm\\
\midrule
DreamTIP-DWL & $31.89 \pm 0.61$ & $28.20 \pm 0.90$ & $20.42 \pm 1.23$ & $17.30 \pm 2.62$ & $36.06 \pm 0.87$ & $35.40 \pm 0.55$ & $20.92 \pm 1.24$ & $6.82 \pm 0.43$ \\ 
DreamTIP-DeepSeekV3 & \textbf{32.15 $\pm$ 0.21} & \textbf{31.06 $\pm$ 0.55} & $20.74 \pm 1.52$ & $17.89 \pm 1.66$ & $36.13 \pm 0.28$ & $35.46 \pm 0.13$ & $25.79 \pm 1.29$ & $7.57 \pm 0.88$ \\ 
DreamTIP-GPT5 & $31.21 \pm 1.11$ & $30.55 \pm 0.93$ & \textbf{22.40 $\pm$ 1.70} & \textbf{18.56 $\pm$ 1.64} & \textbf{36.47 $\pm$ 0.48} & \textbf{35.53 $\pm$ 0.83} & \textbf{31.71 $\pm$ 1.11} & \textbf{8.97 $\pm$ 1.07} \\ 
\bottomrule
\end{tabular}
\caption{Performance comparison of different Task-Invariant Properties design methods in simulation. \textbf{Bolded} numbers indicate the best performance.}
\label{tab:ablationtable}
\end{table*}
% As shown in Fig.~\ref{fig:Main results} 展示了上述方法在八个迁移任务上的性能表现，纵轴表示轨迹平均奖励，横轴表示任务的不同难度，我们用随机3个种子的超过100条轨迹测试得到。
%图x说明，在任务难度较低时，各方法的性能表现相差不大，而随着任务难度不断提高，对模型的适应能力提出了更高的要求，我们的方法表现出了最强的鲁棒性。随着难度增加，其性能下降的幅度是最小的，我们的方法在几乎所有任务的性能表现方面显著高于基线方法，例如在26—23cm的Crawl的任务上，基线方法在最初的难度上平均奖励约为33.51，而在最高难度上平均奖励下降至5.66，性能下降幅度约为83.1%，与之相比，我们的方法从36.58来到25.35，在baseline在该任务上几乎失效的情况下，性能下降幅度仅为30.6%。
%在Stair，Gap，Climb，与Compound Transsfer任务上，Ours w/o TIP方法在这些任务上的表现都明显好于缺少正则化约束的WMP w/o fine-tune方法，而后者在COmpound Transfer上的gap与stair任务上性能表现甚至弱于微调前，这说明加入了正则化约束的方法有效防止了微调过程中的表征漂移和知识遗忘，始终保持着更强的适应性能。
%从基线模型与 Ours w/o Adapt 的对比结果可以看出，引入任务不变性属性作为辅助预测目标后，世界模型的泛化性能显著增强。在几乎所有迁移任务上，该方法均优于基线模型，验证了任务不变性属性在提升迁移泛化能力方面的有效性。
As illustrated in Fig.~\ref{fig:Main results}, the performance of the aforementioned methods across eight transfer tasks is evaluated. Our method achieves an average performance improvement of 28.1\% across eight distinct simulated transfer tasks. The figure shows that performance differences are small under low task difficulty but become pronounced as difficulty increases, demanding greater adaptability. Our method proves most robust, with the least performance degradation as tasks grow harder. It consistently outperforms all baseline methods in nearly every task. For example, in the Crawl task with gap widths varying from 26 \textit{cm} to 23 \textit{cm}, the WMP method achieves an average reward of approximately 33.51 at the easiest level, which sharply decreases to 5.66 at the highest difficulty level, corresponding to a performance drop of about 83.1\%. In contrast, our method declines from 36.58 to 25.35, resulting in a significantly more moderate reduction of only 30.6\%, even under conditions where the baseline approach almost completely fails.

On tasks such as Stair, Gap, Climb, and Compound Transfer, the proposed method without Task-Invariant Properties (\textbf{Ours w/o TIP}) consistently outperforms the Weighted Model Predictive Control with Finetuning (\textbf{WMP w/ Finetune}) approach, which lacks explicit regularization constraints. Notably, the latter exhibits even worse performance on gap and stair tasks within the Compound Transfer setting compared to its pre-finetuned version. These results suggest that the integration of regularization constraints effectively mitigates issues such as representation drift and knowledge forgetting during the adaptation process, thereby enabling more robust and stable performance across tasks.

Furthermore, comparisons between the baseline (\textbf{WMP}) and our method without adaptation mechanisms (\textbf{Ours w/o Adapt}) reveal that learning with Task-Invariant Properties significantly enhances the task transfer capacity of the world model. The proposed approach surpasses the baseline in almost all transfer tasks, confirming the effectiveness of leveraging Task-Invariant Properties to improve transfer capability. This design allows the model to better capture structural and dynamic features that remain consistent across environments and tasks, thereby maintaining reliable prediction and decision-making capabilities even when confronted with unseen or highly challenging scenarios.

%As shown in Fig.~\ref{fig:comparison}，在crawl任务中，我们分别在仿真环境和真实环境中对比了基线方法和本文方法的实验结果。由图可知，当机器狗爬行通过障碍物时，基线方法存在碰撞的情况，而本文方法则能无碰撞通过。
% As shown in Fig.~\ref{fig:comparison}, a comparative evaluation between the proposed method and the baseline is conducted in both simulated and real-world crawl tasks, specifically involving traversal over a 25 \textit{cm} high obstacle. The results clearly demonstrate the critical advantage of our approach sim-to-real transfer capability and robustness. In the simulation environment (above the gray dashed line), the baseline method fails to safely navigate the obstacle. The yellow dots, which mark the robot’s traversal height, clearly indicate that its head exceeds the red obstacle line, resulting in a collision annotated as “Hit the obstacle.” The same failure is consistently observed in the real-world setup (below the dashed line), where the robot physically collides with the obstacle due to incorrect body posture planning during traversal. In contrast, our method consistently achieves safe passage in both environments. The yellow dots confirm that the robot maintains height above the red obstacle line, successfully avoiding contact under both simulated and physical conditions. This consistent performance, combined with the stark contrast to the baseline’s failure, convincingly demonstrates the capability of our method to effectively bridge the simulation-to-reality gap.

\subsection{Real-World Evaluation}
\begin{figure}[t]
    \centering
    \includegraphics[width=0.68\columnwidth, trim=580 90 480 170, clip]{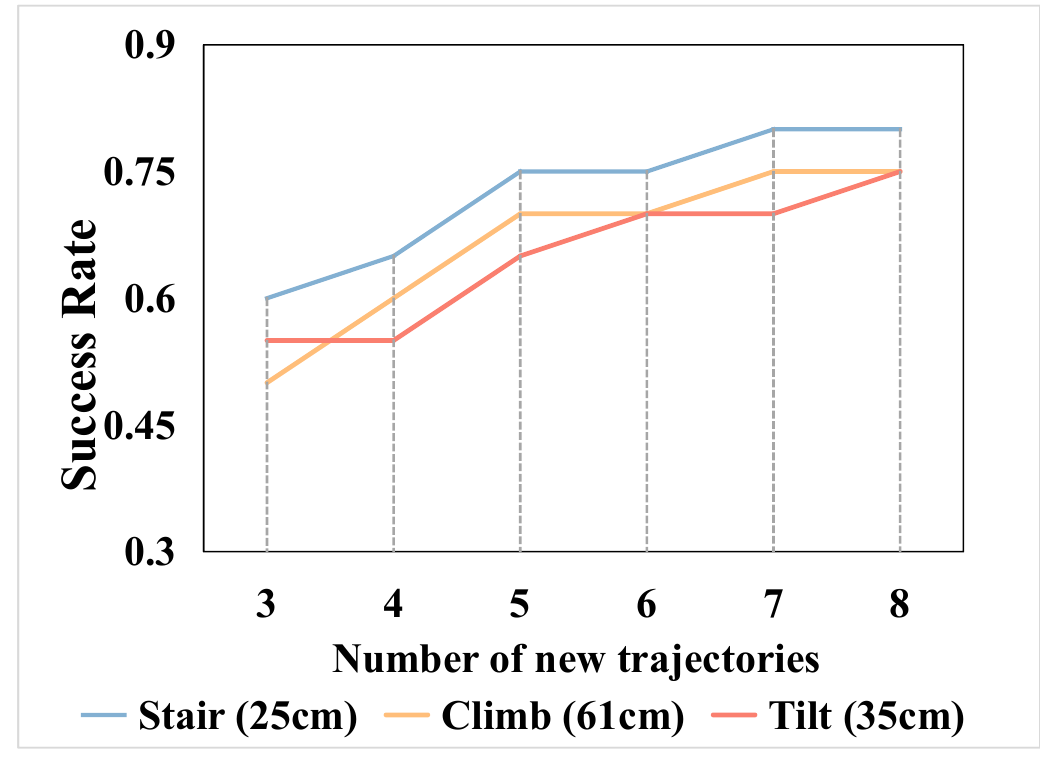}
    \caption{Ablation on the number $n$ of trajectories added to the mix buffer. We evaluated the performance of our approach in a simulation environment on the Stair (25 \textit{cm}), Climb (61 \textit{cm}), and Tilt (35 \textit{cm}) tasks, collecting additional trajectories on these transfer tasks for adaptation after pre-training DreamTIP. Success rates are calculated over twenty trials.}
    \label{fig:traj ablation}
\end{figure}

% 我们将所提方法（Ours）、Ours w/o Adapt以及基线模型（WMP）部署于真实的 Go2 机器人上，并在四种不同难度的地形上进行了综合性能评估，包括：stair(16cm), climb(52cm), tilt(33cm), crawl(25cm)，以成功率（10条轨迹统计）作为评价指标。为测试动力学变化下的迁移能力，我们在机器人基座右侧加装2kg配重模拟质心偏移。所有测试均以0.6 m/s的速度指令执行以确保运动一致性。
%实验结果如Tab. \ref{tab:Real World Evaluation}所示。实验表明，基线方法在 stair 任务中表现良好，在 crawl 任务中也较为稳定，但在 climb 和 tilt 任务中与本文方法差距显著。相比之下，本文方法展现出更强的跨任务迁移能力，尤其在动态性强、对动力学变化敏感的 climb（52cm）任务中优势突出：基线成功率仅为10%，而本文无适应版本（Ours w/o Adapt）达90%，完整方法（Ours）达到100%。结果表明，任务不变性属性学习与适应更新机制有效缓解了仿真与现实差异，在复杂动态任务中表现出优越的迁移能力。
We deployed our method (\textbf{Ours}), its non-adaptive variant (\textbf{Ours w/o Adapt}), and the baseline (\textbf{WMP}) on a Unitree Go2 robot, evaluating their performance across four terrains: Stair (16 \textit{cm}), Climb (52 \textit{cm}), Tilt (33 \textit{cm}), and Crawl (25 \textit{cm}). Success rates were computed over 10 trials per task. To test robustness under dynamic changes, a 2 $kg$ counterweight was attached to the robot’s right side, shifting its center of mass. All tests used a velocity command of 0.6 $m/s$ to ensure consistent motion.

The results presented in Tab.~\ref{tab:Real World Evaluation} indicate that while the baseline method performed competently in Stair and Crawl tasks, it exhibited substantially inferior performance in Climb and Tilt scenarios compared to our proposed approach. Our method exhibits stronger transfer capability, especially in the 52 \textit{cm} Climb task: the baseline achieved a mere 10\% success rate, whereas the ablated version without adaptation (\textbf{Ours w/o Adapt}) and the full method (\textbf{Ours}) attained success rates of 90\% and 100\%, respectively. 
% \color{red}
The baseline method exhibited lower performance on the Go2 platform compared to Go1 \cite{lai2025world}, likely due to Go2's greater mass, larger size, and consequently higher control difficulty and sim-to-real transfer requirements. In contrast, our approach demonstrates stronger adaptability.
% \color{black}
The results demonstrate that our method effectively narrows the sim-to-real gap and enables robust policy transfer under challenging real-world conditions, confirming its capacity to handle dynamic environmental variations while maintaining strong cross-task adaptability.

As shown in Fig.~\ref{fig:comparison}, the proposed method and the baseline are compared in both simulation and real-world crawl tasks. The results demonstrate that while the baseline causes collisions when the robot dog traverses obstacles, our method achieves a safe pass.

\subsection{Ablation Study}
%任务鲁棒性特征的设计
% 我们研究了三种任务不变性属性的设计方式对模型性能的影响，在仿真环境中通过 Climb 和 Tilt 两个迁移任务比较了各方法的平均轨迹奖励。具体方法包括：DreamTIP-DWL（直接预测特权信息）、DreamTIP-GPT-5（本文主要采用的大语言模型用于提取任务不变性属性）以及用于对比的 DreamTIP-DeepSeekV3（基于另一大语言模型构建属性提取函数）。如 Tab.~\ref{tab:Real World Evaluation} 所示，基于大语言模型构建任务不变性属性作为辅助预测目标的方法，在两项任务中的性能均显著优于直接预测特权信息的 DreamTIP-DWL。这表明LLMs设计的任务不变性属性有效捕捉到了能够促进世界模型迁移性能提升的关键
We evaluate three designs of Task-Invariant Properties in the Climb and Tilt transfer tasks under simulation, using average trajectory reward as the metric. The methods include: \textbf{DreamTIP-DWL} (predicting privileged information directly), \textbf{DreamTIP-GPT5} (our main LLM-driven properties designed method), and \textbf{DreamTIP-DeepSeekV3} (another LLM for comparison). As Tab.~\ref{tab:Real World Evaluation} shows, LLM-based methods that construct Task-Invariant Properties as auxiliary targets significantly outperform direct privileged information prediction.
% \color{red}
This result represents the best outcome from three independent TIP generations produced by different LLMs,
% \color{black}
indicating that the properties designed by LLMs effectively capture essential features that enhance the transfer performance of the world model.

%微调时新收集的轨迹数量n影响着微调的实际性能，一般来说收集轨迹越多，模型的性能也会相应上升，因为其覆盖了更多的现实动力学分布情况，但总所周知在真实环境中收集轨迹是十分昂贵的，因此我们需要在其中做出trade off。为了探究轨迹数量对微调效果的实际影响，使用ours方法vary n from 3 to 8。如图x所示，我们在仿真环境中的三个迁移任务上，stair(25cm),climb(61cm),Tilt(35cm),分别探究了微调时使用不同轨迹数量在对应迁移任务上的性能表现，直观来说，轨迹数量从3提升到5时，性能上升较为明显，而在轨迹数量到达5之后，性能的上升幅度不高的同时收集数据的成本变大，因此我们在实际实验开展时选择收集5条轨迹进行微调
The number of trajectories $n$ used during fine-tuning significantly affects model performance. While more data generally improves robustness by covering broader real-world dynamics, collecting such data is costly. Using our method, we varied $n$ from 3 to 8 and evaluated performance on three simulated transfer tasks: Stair (25 \textit{cm}), Climb (61 \textit{cm}), and Tilt (35 \textit{cm}). As shown in Fig. 6, performance improves noticeably when the number of trajectories increases from 3 to 5, but exhibits diminishing returns beyond this point. We therefore set $n=5$ in practice to balance effectiveness and computational cost.

\section{CONCLUSIONS}
%本文提出了 DreamTIP 框架，它在Dreamer的基础上引入了任务不变性属性学习，以提升四足机器人运动任务中的虚实迁移能力。通过借助大语言模型（LLM）指导提取诸如接触稳定性、地形间隙等对动力学鲁棒且跨任务可迁移的特征，DreamTIP 能够学习到对环境变化不敏感的潜在动力学表征。此外，为了进一步缓解仿真与真实之间的差异，我们设计了一种高效的适应策略，结合混合回放缓冲池与正则化约束，使 DreamTIP 能够在避免表征崩塌和灾难性遗忘的同时，快速校准至真实动力学分布。在复杂地形（如楼梯、跳跃和攀爬场景）上的广泛实验表明，DreamTIP 在仿真和真实环境中均显著优于强基线方法。尤其是在最困难的 Crawl 任务（23 cm）中，DreamTIP 的平均奖励达到 25.35，远超基线方法的 5.66；在 Unitree Go2 真实机器人平台的 Climb 任务中，DreamTIP 达到 100% 的成功率，而基线仅为 10%。这些结果充分验证了在世界模型中引入任务不变性属性能够显著增强策略的泛化能力，并有效弥合虚实之间的差距。未来的工作中，我们计划探索在更多样化的仿真与真实数据混合下训练世界模型，并扩展 DreamTIP 至更多任务。我们相信，这一研究方向为实现鲁棒、可扩展且具备自然适应能力的机器人学习提供了有前景的路径。

This paper introduces DreamTIP, an extension of the Dreamer framework designed to improve sim-to-real transfer in quadruped robot locomotion through Task-Invariant Properties learning. By leveraging large language models, DreamTIP learns dynamics-robust and task-invariant properties , such as contact stability and terrain clearance, to reduce the reliance on specific dynamic parameters. To further narrow the sim-to-real gap, we propose an efficient adaptation strategy integrating a mix buffer with regularization constraints, which enables stable calibration to real-world dynamics while alleviating representation collapse and catastrophic forgetting. Extensive evaluations across various transfer tasks show that DreamTIP consistently outperforms baselines in both simulated and real-world settings. These results underscore the value of Task-Invariant Properties in enhancing policy generalization and sim-to-real transfer. 
% \color{red}
However, this work still has some limitations. For example, prolonged operation leads to a certain degree of performance degradation due to the compounding errors in the world model.
Future work will explore leveraging richer simulated and real-world data to improve the world model's long-term prediction accuracy and robustness, thereby mitigating performance degradation from error accumulation and
% \color{black}
offering a robust and scalable framework for adaptive robot learning.


%%%%%%%%%%%%%%%%%%%%%%%%%%%%%%%%%%%%%%%%%%%%%%%%%%%%%%%%%%%%%%%%%%%%%%%%%%%%%%%%

%%%%%%%%%%%%%%%%%%%%%%%%%%%%%%%%%%%%%%%%%%%%%%%%%%%%%%%%%%%%%%%%%%%%%%%%%%%%%%%%

%%%%%%%%%%%%%%%%%%%%%%%%%%%%%%%%%%%%%%%%%%%%%%%%%%%%%%%%%%%%%%%%%%%%%%%%%%%%%%%%

%%%%%%%%%%%%%%%%%%%%%%%%%%%%%%%%%%%%%%%%%%%%%%%%%%%%%%%%%%%%%%%%%%%%%%%%%%%%%%%%

\bibliographystyle{IEEEtran}
\bibliography{ref}

\end{document}